\documentclass[journal]{IEEEtran}

\usepackage{mathrsfs}
\usepackage{amsmath,amsfonts}
\usepackage{array}
\usepackage{bm}
\usepackage[caption=false,font=normalsize,labelfont=sf,textfont=sf]{subfig}
\usepackage{atbegshi}
\usepackage{stfloats}
\usepackage{url}
\usepackage{verbatim}
\usepackage{subcaption}
\usepackage{subfig}
\usepackage{subfloat}
\usepackage{mathtools}
\usepackage{amsmath}
\usepackage{times}
\usepackage{balance}
\usepackage{graphicx}
\usepackage{amssymb}  	
\usepackage{dsfont}
\usepackage{mathrsfs}
\usepackage{algorithmic}
\usepackage{algorithm}
\usepackage[dvipsnames]{xcolor}
\usepackage[numbers]{natbib}
\usepackage{amsmath,amssymb,amsfonts}
\usepackage{multirow}
\usepackage{graphicx}
\usepackage{rotating}
\usepackage{multicol}
\usepackage{amsmath}

\usepackage{float}
\usepackage{caption}
\usepackage{subcaption}
\usepackage[dvipsnames]{xcolor}
\usepackage{}
\usepackage{graphicx}
\usepackage{color, soul, colortbl}
\usepackage{nicematrix}
\usepackage{colortbl} 
\usepackage{listings}
\usepackage{tabularray}
\usepackage{array}
\usepackage{makecell}
\usepackage{hyperref}
\usepackage{alltt}
\usepackage{adjustbox}  

\usepackage{booktabs}
\usepackage{adjustbox}

\usepackage{tabularx}  
\usepackage{xcolor}

\def\BibTeX{{\rm B\kern-.05em{\sc i\kern-.025em b}\kern-.08em
    T\kern-.1667em\lower.7ex\hbox{E}\kern-.125emX}}
\usepackage{balance}
\begin{document}
\title{
LLM-Guided Task and Motion Planning using Knowledge-based Reasoning
}
\author{Muhayy Ud Din$^{1}$, Jan Rosell$^{2}$,  Waseem Akram$^{1}$, Isiah Zaplana$^{2}$, Maximo A Roa$^{3}$,  and Irfan Hussain$^{1,*}$
\thanks{$^{1}$ Khalifa University Center for Autonomous Robotic Systems (KUCARS), Khalifa University, United Arab Emirates.}%
 \thanks{$^{2}$ Institute of Industrial and Control Engineering (IOC), Universitat Politècnica de Catalunya, Spain.}
 \thanks{$^{3}$ Institute of Robotics and Mechatronics, German Aerospace Center (DLR), Germany.}
\thanks{$^{*}$ This publication is based upon work supported by the Khalifa University of Science and Technology under Award No. RC1-2018-KUCARS, and by the European Commission’s Horizon Europe Framework Programme with the project IntelliMan (AI-Powered Manipulation System for Advanced Robotic Service, Manufacturing and Prosthetics) under Grant Agreement 101070136.\\ 
 Corresponding Author, Email: irfan.hussain@ku.ac.ae}
}





\maketitle

\begin{abstract}
Performing complex manipulation tasks in dynamic environments requires efficient Task and Motion Planning (TAMP) approaches that combine high-level symbolic plans with low-level motion control. Advances in Large Language Models (LLMs), such as GPT-4, are transforming task planning by offering natural language as an intuitive and flexible way to describe tasks, generate symbolic plans, and reason. However, the effectiveness of LLM-based TAMP approaches is limited due to static and template-based prompting, which limits adaptability to dynamic environments and complex task contexts. To address these limitations, this work proposes a novel \hbox{Onto-LLM-TAMP} framework that employs knowledge-based reasoning to refine and expand user prompts with task-contextual reasoning and knowledge-based environment state descriptions. Integrating domain-specific knowledge into the prompt ensures semantically accurate and context-aware task plans. The proposed framework demonstrates its effectiveness by resolving semantic errors in symbolic plan generation, such as maintaining logical temporal goal ordering in scenarios involving hierarchical object placement. 
The proposed framework is validated through both simulation and real-world scenarios, demonstrating significant improvements over the baseline approach in terms of adaptability to dynamic environments and the generation of semantically correct task plans. Videos and code can be found here: \textcolor{blue}{~\url{https://muhayyuddin.github.io/llm-tamp/}}.

\end{abstract}

\section{Introduction}

Task and motion planning (TAMP) is an essential component in sequential manipulation to perform tasks in complex environments. TAMP involves two stages: high-level symbolic plan generation, which develops a long-term abstract action sequence, and low-level motion planning, which determines the trajectories with geometric constraints required to execute the computed symbolic plan. The classical solution to TAMP problems consists of describing the task in a specific planning representation, such as the Planning Domain Description Language (PDDL)~\cite{aeronautiques1998pddl}, and using a task planner, such as GraphPlan~\cite{blum1997fast} or FastForward~\cite{hoffmann2001ff}, to compute a symbolic plan, whose execution requires the call to a motion planner for all actions involving movements. A custom-designed communication layer is usually used to enable interaction between the task and motion planning modules.


Describing the task specification using PDDL for TAMP has proved to be very successful. However, PDDL has some drawbacks; it requires manual coding of all real-world descriptions and task constraints, which is impractical in a dynamic environment, and in addition, its syntax may be difficult for new users to understand~\cite{chen2023autotamp}. On the other hand, compared to PDDL, natural language offers a more intuitive and user-friendly interface for task descriptions~\cite{ding2023integrating}. In this line, recent advances in pre-trained LLMs, such as GPT-4~\cite{achiam2023gpt}, have shown good performance in various natural language tasks, and this has accelerated research on the use of LLMs in task planning~\cite{Singh2022} and TAMP~\cite{ding2023task}~\cite{Wang2024}.


LLM-based methods for TAMP typically use pre-trained LLM models, such as GPT-4, that receive task descriptions as input~\cite{Wang2024}.  The LLM module is responsible for computing symbolic plans and reasoning for failures, showing significant capabilities when in-context prompting is provided~\cite{dong2022survey}. However, a weak prompt could drastically reduce planning and reasoning capabilities, and hence several recent studies proposed different ways of prompt engineering for better task elaboration. For example, a template-based prompt is proposed in~\cite{Wang2024}, where fixed templates with variables are used to describe spatial and geometric relations, or the proposal in~\cite{ding2023task}, where a template is used to describe the scene and another to textualize the state of the environment. To better exploit natural language, an interactive task planning using LLM was proposed in~\cite{Li2023}, where the user input was integrated with stored task guidelines to generate the prompt.


Most of the above-stated prompt generation methods rely on fixed templates.  This can result in static and repetitive prompts that may not effectively capture the dynamic nature of the environment, i.e., these approaches may not generalize well to new or unexpected situations, thus leading to suboptimal prompts when the task context changes.  In addition, fixed templates may not capture the detailed context of a task, as they often oversimplify the complex states of the environment. This might result in incomplete or misleading prompts, which may lead to misunderstandings in task planning. Therefore, since effective prompting methods are crucial to improve LLM performance~\cite{Jin2024}, robust and refined prompt engineering techniques are needed to support the elaboration of accurate task plans.


To address the above-stated challenges, this study proposes an  Onto-LLM-TAMP approach that utilizes knowledge-based reasoning to refine and expand the user’s prompt within the task context. Building on the baseline LLM3 TAMP framework~\cite{Wang2024}, we extend it with our novel Onto-LLM-TAMP method. 
Some other approaches, such as \cite{Li2024}, have also proposed to integrate ontological knowledge with LLM-based task planning, but their primary focus is on object properties to define optimal grasping strategies (e.g., \texttt{glass} $\rightarrow$ \texttt{pick gently}, \texttt{wood} $\rightarrow$ \texttt{pick stably}). On the contrary, we employ knowledge-based reasoning for prompt tuning to refine the action sequence for correct temporal goal ordering, integrating it with the user's prompt input for the LLM. This ensures that the generated plan is not only correct but also semantically accurate. For example, consider the scenario 
, where the user instructs to \textit{\small"put the banana, apple, and bowl in the plate"}. In most cases, the symbolic baseline planner generates a sequence like \textit{\small"pick banana and apple, and place in the plate"} followed by \textit{\small" pick bowl and place in the plate"}. Although it is syntactically correct, it is semantically flawed, as it could result in placing the bowl on top of the apple and banana. Our proposed Onto-LLM-TAMP approach addresses this by incorporating domain-specific knowledge. It integrates a guiding note into the prompt, specifying that 
\textit{\small"put bowl before food items because crockery has priority over food items. Put banana and apple after crockery because food items have less priority."}  This extended prompt will help the LLM to generate a semantically correct and consistent plan. \\ \\
\textit{\textbf{Contributions}}\\
The core contribution of this work is a novel  Onto-LLM-TAMP framework for task and motion planning that helps LLM to compute semantically accurate plans. Building upon this core framework, the following contributions are highlighted:
\begin{itemize}
    \item \textit{Automated prompt-tuning with task contextual reasoning:} This approach extracts task-related common sense and procedural knowledge from the ontology to guide the LLM to generate a semantically accurate and logically ordered plan.

    \item \textit{Knowledge-based environment state description:} We utilize ontological knowledge to construct detailed, context-aware descriptions of the environment, enhancing the prompt-tuning process for more effective task planning.

\item \textit{Benchmarking of LLM models:} We evaluated several major LLM models, including GPT, LLaMA, Gemini, and Cohere, in both standard LLM-TAMP and our Onto-LLM-TAMP frameworks to assess the impact of prompt tuning on the LLM performance.
\end{itemize}
The rest of the paper is structured as follows: Sec.~\ref{sect:related} provides a detailed overview of the current state-of-the-art methods in Task and Motion Planning, highlighting the role of Large Language Models. Sec.~\ref{sec:pf} formulates the problem statement and presents the proposed Onto-LLM-TAMP framework. Sec.~\ref{sec:approach} describes the proposed approach; elaborates on the methodology, and provides details of the ontology integration, and the execution pipeline. Sec.~\ref{resultndiscussion} presents experimental results and analysis, demonstrating the framework’s effectiveness in simulated and real-world environments. Finally, Sec.~\ref{conclusion} concludes the work.

\section{Related Work}\label{sect:related}

\subsection{Task and Motion Planning}


TAMP uses a structured approach where task planning and motion planning are handled separately but in a coordinated manner. Task planning typically uses symbolic AI techniques, such as hierarchical planning, to generate a sequence of discrete actions based on predefined rules or logic. Motion planning uses geometric algorithms, such as sampling-based planners, to compute feasible paths~\cite{orthey2023sampling}~\cite{rosell2019planning}. A declarative language, such as PDDL, is used for TAMP. However, its use is limited due to the requirement of a large number of parameters to define the problem, actions, and effects \cite{jiang2019task}. Moreover, as the action space expands, finding geometrically feasible symbolic action sequences becomes computationally challenging without effective heuristics~\cite{garrett2020pddlstream}. Recent studies have explored data-driven heuristics to enhance TAMP efficiency~\cite{noseworthy2021active}~\cite{yang2022sequence}. 
Other approaches use knowledge-based reasoning to reduce the action space, using domain-specific ontological knowledge, and then apply heuristics for efficient symbolic plan generation~\cite{akbari2016task}~\cite{akbari2015reasoning}. 
The approach proposed here is a general manipulation framework to work across various domains, with the specific application ontology providing the specialized knowledge needed to correctly solve the task.

\subsection{Large Language Models}
Large models represent a significant advancement in artificial intelligence, enabling powerful language understanding and general reasoning capabilities in various domains. Several models have been proposed including GPT-4~\cite{openai2023gpt4}, Gemini~\cite{anil2023gemini}, LLaMA~\cite{touvron2023llama}, Cohere Language Models~\cite{cohere2023}, and DeepSeek~\cite{xu2023deepseek}. These models are built on transformer architectures trained on vast datasets. GPT-4 excels in complex reasoning and code synthesis, while Gemini integrates multimodal capabilities, allowing it to process text, images, and video. LLaMA stands out as an efficient open-source model that facilitates research and development by the broader community. Cohere offers industry-focused models that are optimized for industry-related applications. DeepSeek is another emerging model that aims to bridge the performance gap between open-source models and cutting-edge proprietary systems, focusing on multilingual and domain-specific tasks.


\subsection{LLMs for TAMP}
These foundation models are increasingly being used in task planning and symbolic reasoning for robotics and automated systems, i.e., LLMs are being explored to enhance task planning by using their ability to understand and generate human-like instructions, descriptions, and planning sequences. In particular, they allow to translate natural language task descriptions into formal task representations without the need to manually represent the problem domain~\cite{huang2022language,brown2020language}. Due to their extensive language understanding and contextual reasoning capabilities, LLMs are used to generate high-level task plans and their translation into feasible motion sequences with minimal additional training~\cite{ahn2022can,huang2023grounded}.

For example, GPT-4 has been adapted for robotic task planning by generating high-level action sequences and providing natural language explanations for task execution \cite{vemprala2023chatgpt}. Google's SayCan system combines LLMs with affordance-based planning to ground language commands into feasible robot actions \cite{ahn2022saycan}. Similarly, language models have been explored as zero-shot planners, enabling agents to extract actionable plans directly from textual instructions \cite{huang2022languageplanning}. These approaches demonstrate the potential of foundation models to serve as cognitive reasoning engines in task and motion planning (TAMP).
For instance, LLM-GROP~\cite{ding2023task} utilizes a pre-trained LLM to define symbolic goals for object placement and rearrangement, integrating these with traditional TAMP methods, and AutoTAMP~\cite{chen2023autotamp} employs a pre-trained LLM to convert user inputs into formal language, which is then used in a TAMP algorithm.

In all the approaches mentioned above, effective prompting techniques are crucial to optimize LLM performance~\cite{Jin2024}. In addition,~\cite{vemprala2023chatgpt} identified key challenges in prompting LLMs for robotic manipulation, including the need for detailed and accurate problem descriptions, as well as managing biases that shape response structures. This study tackles the issue of effective prompting for task and motion planning by utilizing ontological knowledge to enhance task elaboration for LLM.

\begin{figure*}
\centering
\includegraphics[width=\linewidth]{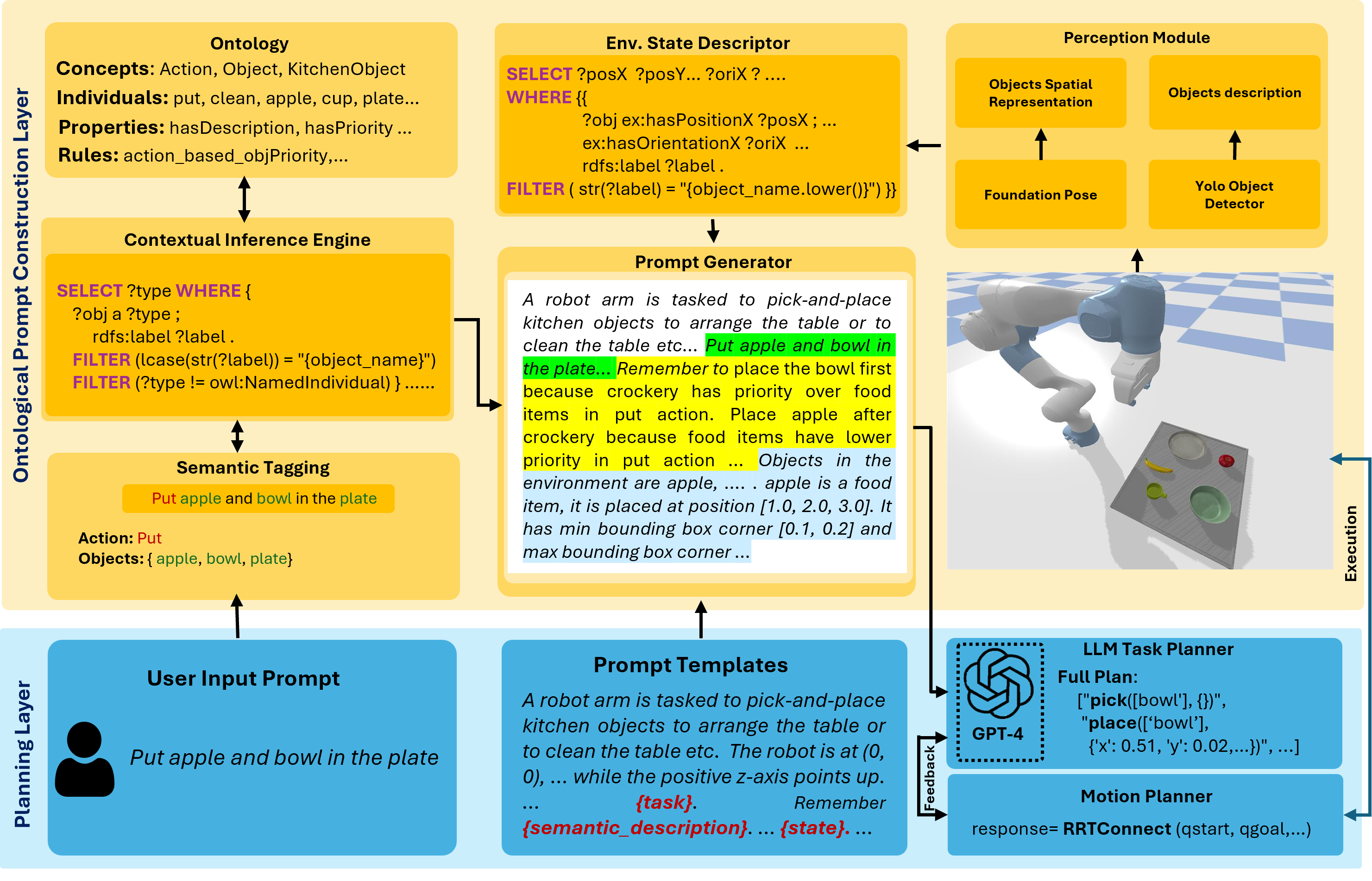}
\caption{The Onto-LLM-TAMP framework enhances prompt elaboration for generating semantically accurate symbolic plans. It begins by processing the user input to extract actions and objects through semantic tagging. The Contextual Inference Engine uses SPARQL queries to retrieve object types and priorities from the ontology, ensuring the correct action sequence based on predefined rules. The Perception Module, with YOLO-based object detection and FoundationPose for object pose estimation, provides real-time spatial data. This information is textualized using ontological knowledge by the Environmental State Descriptor and fed into the Prompt Generator. The final prompt is then fed into the LLM Task Planner, which produces a structured task plan. Finally, the Motion Planner ensures the robot executes the task with feasible, collision-free movements.}
\label{fig:framework}
\end{figure*}

\section{Problem Formulation}\label{sec:pf}
This section formally defines the task and motion planning problem and how we model the problem for Onto-LLM-TAMP.
\subsection{Task and Motion Planning}
A task and motion planning problem is typically described as a tuple \(\mathcal{X} = \langle\mathcal{O}, \mathcal{S}, \mathcal{A}, \mathcal{T}, s_0, g\rangle \), where: 
\begin{itemize} 
\item $\mathcal{O}$ denotes the set of objects in the environment. 
\item $\mathcal{S}$ represents the state space of all possible configurations of objects within the environment, where $s_t \in \mathcal{S}$ signifies the state of the objects at time $t$. 
\item $\mathcal{A}$ is the set of primitive actions, each one requiring the call to the motion planner for its execution. A primitive action $a \in \mathcal{A}$ is defined by a set of objects $o\in \mathcal{O}$ and a set of continuous parameters $\gamma$, e.g., \texttt{Place(cup, [x, y, z, $\theta$])}. The parameters $\gamma$ serve as goals for the motion planner and $a$ is considered feasible if the motion planner finds a solution, i.e., a collision-free trajectory $\tau$. We denote such a feasible action as $a(\tau)$. 
\item $\mathcal{T}$ is the state transition function that generates the subsequent state $s_{t+1}$ after executing an action $a_t(\tau_t)$ at state $s_t$, formally represented as $s_{t+1} = \mathcal{T}(s_t, a_t(\tau_t))$. This transition function can be evaluated with a black-box simulator. 
\item $s_0$ is the initial state, with $s_0 \in \mathcal{S}$.
\item $g$ is the goal function, defined as $g(s) : \mathcal{S} \rightarrow \{0, 1\}$, which checks if the task goal has been met at a given state $s$. 
\end{itemize}

The objective in Task and Motion Planning is to derive a sequence of feasible actions $\mathbf{a} = \{a_0(\tau_0), a_1(\tau_1), \dots, a_T(\tau_T)\}$, such that $s_{t+1} = \mathcal{T}(s_t, a_t(\tau_t))$, for each $t = 0, 1, \dots, T$, and $g(s_{T+1}) = 1$.

\subsection{Problem Modeling}
The Onto-LLM-TAMP framework is defined as a tuple \(\langle \mathcal{P}, \mathcal{K}, \xi, \mathcal{X}, \mathcal{L} \rangle\), where each component plays a crucial role in structuring a feasible action plan:
\begin{itemize}
    \item \(\mathcal{P}\) represents the prompt, containing essential information such as the task description, environment state, and details about the objects involved.
    \item \(\mathcal{K}\) denotes the ontological knowledge-base, which provides domain-specific information and task-related rules.
    \item \(\xi\) is the knowledge reasoner, responsible for enriching the prompt with task-specific correctness knowledge:    
    \begin{equation}
    \xi : \mathcal{P} \times \mathcal{K} \to \epsilon
    \end{equation}
where \(\epsilon\) represents additional guidance relevant to the task.
    \item \(\mathcal{X}\) is the set of task parameters.
    \item \(\mathcal{L}\) is the LLM model, such as GPT-4, that operates as a black-box component, defined by:
    \begin{equation}
    \mathcal{L} : \mathcal{P} \times \epsilon \times \mathcal{X} \rightarrow \mathbf{a}
    \end{equation}
\end{itemize}
This model receives the prompt \(\mathcal{P}\), enhanced task knowledge \(\epsilon\), and TAMP parameters \(\mathcal{X}\), and it outputs a set of feasible actions \(\mathbf{a} \subset \mathcal{A}\) that enable the robot to transition from the initial environment state to the desired goal state. 

The effectiveness of \(\mathcal{L}\) depends significantly on the quality of the prompt, since an elaborate and semantically accurate prompt improves the probability of generating a valid sequence of actions. The primary objective of this work is to explore how ontological knowledge and semantic reasoning can further refine and enrich prompts, thereby enhancing the accuracy of semantically correct action sequence generation for consistent task execution. 



\section{Proposed approach}\label{sec:approach}
This section explains how ontological knowledge is used to generate suggestions for the LLM, ensuring semantically correct symbolic plan generation with proper temporal goal ordering.
The architecture of our Onto-LLM-TAMP framework is illustrated in Fig.~\ref{fig:framework}. It is composed of \textit{Ontological Prompt Construction Layer} and the \textit{Planning Layer}.

The \textit{Ontological Prompt Construction Layer} is responsible for creating a prompt to be sent to the LLM-based planning module. This is achieved by using user input, predefined prompt templates, and the information available in an ontology. It comprises six modules:
   1) the  \textit{Ontology} module that contains the application ontology, comprising classes, properties, and rules (detailed in Sec.~\ref{sec:ontology});
   2) the  \textit{Semantic Tagging} module that analyzes the user input and facilitates the queries to the ontology (discussed in Sec.~\ref{sec:tagging});
   3) the  \textit{Contextual Inference Engine} module that, using inferred knowledge composed of priorities and rules based on the relationships between objects and actions, suggests the right order of actions in the plan to perform the task in a semantically and logically correct manner (explained in Sec.~\ref{sec-inference});
   4) the \textit{Perception Module} that identifies all objects within the environment, estimating their spatial positions and orientations using YOLO~\cite{Jocher_Ultralytics_YOLO_2023} and FoundationPose~\cite{wen2024foundationpose}, and instantiating them in the ontology, as done in a similar way in~\cite{ruiz-celada2024_situationAwareness};
   5) the \textit{Environmental State Descriptor} module that translates the environmental state retrieved from the ontology into a text format, describing the spatial layout and types of objects in the environment (the perception module and the environment state descriptor are detailed in Sec.~\ref{sec-env});
   6) The \textit{Prompt Generator} module, that combines the prompt templates with the information coming from the \textit{Contextual Inference Engine} and  \textit{Environmental State Descriptor} modules (explained in Sec.~\ref{sec-promptgen}).

The \textit{Planning Layer} takes a prompt as input and generates a plan to execute. It closely follows the baseline approach~\cite{Wang2024} and comprises three modules: 1): the \textit{User Input} which describes the task to be performed; 2) the \textit{Prompt Template} which contains the static sections of the prompt, provides general task descriptions and structures the overall prompt\footnote{Complete template can be found here: \url{https://github.com/Muhayyuddin/llm-tamp/blob/llm-site/assets/prompt/txt}}; 3) the \textit{LLM-TAMP Module}, detailed in Sec.~\ref{sec-taskplanner}, that has the \textit{LLM Task Planner} to compute a symbolic plan, and a motion planner to compute the motions for each action of the plan.

\subsection{Ontological Knowledge Representation}\label{sec:ontology} 

\begin{figure}
	\includegraphics[width=\columnwidth]{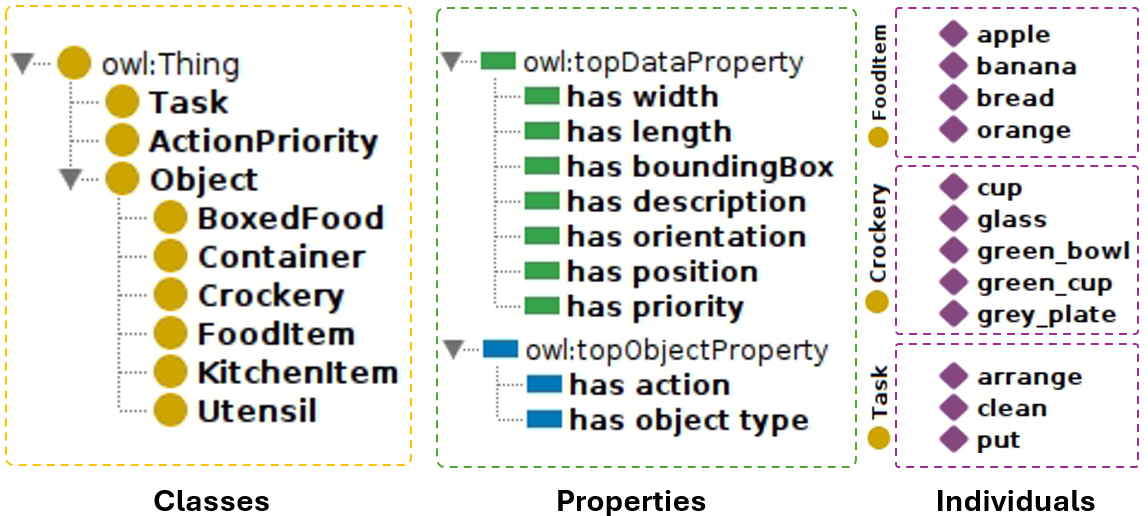}
	\caption{Kitchen ontology, showing the hierarchy of Classes (yellow), Properties (green and blue), and Individuals (purple).}
	\label{fig:ontolist}
\end{figure}
%
We developed a simple kitchen domain ontology to provide semantic reasoning for performing tasks in a kitchen environment. This domain-specific ontology enhances reasoning about task constraints,  priority rules and object relations, and for this, three main classes have been defined: \texttt{Task}, \texttt{ActionPriority} and \texttt{Object}. The  \texttt{Object} class is further subdivided into six subclasses describing different kitchen object types:  \texttt{FoodItems}, \texttt{BoxedFood}, \texttt{Crockery}, \texttt{Utensil},  \texttt{KitchenItems}, and \texttt{Container}. The hierarchy of classes, object properties, data properties, and some example individuals are shown in Fig.~\ref{fig:ontolist}. 

The data properties of the ontology describe the essential attributes of different objects, including action priority values and specific descriptions with temporal constraints, such as \texttt{pick after FoodItem} or \texttt{place Crockery before FoodItem}. These descriptions are utilized by the reasoner for prompt tuning, enabling it to guide action sequencing. Other data properties capture object-specific attributes like position, orientation, bounding box, and additional task-related properties. Object properties, on the other hand, define relationships between objects and actions. For instance, \texttt{has object type} specifies the classification of an object (e.g., FoodItem or Crockery), while \texttt{has action} associates relevant actions, such as ``put," ``clean" or ``arrange" with each object. These properties are essential for knowledge-based reasoning about task execution.

We define a set of rules\footnote{Complete rule set used in this study is available here: \url{https://github.com/Muhayyuddin/llm-tamp/blob/llm-site/assets/smart_kitchen.rdf}} that specify the correct order of action sequences when manipulating objects in the kitchen environment. These rules are implemented as instances of \texttt{ActionPriority}, where each rule associates a task (such as \texttt{put} or \texttt{clean}) with an object type (such as \texttt{FoodItem} or \texttt{Crockery}) and assigns a priority level. Through these rules, the reasoning mechanism infers a semantically correct action sequence for task execution.
For example, Eq.~(\ref{rule1}) and~(\ref{rule2}) show the description logic representation for two rules, named Rule~1 and Rule~2.

 \begin{align}
 \label{rule1}
    \text{\textcolor{Mulberry}{Rule1}} \sqsubseteq \text{\textcolor{orange}{ActionPriority}} \sqcap & \exists \text{\textcolor{ForestGreen}{hasAction}.\textcolor{Mulberry}{Put}}
    \nonumber \\
    \sqcap \exists \text{\textcolor{ForestGreen}{hasObjectType}.\textcolor{orange}{Crockery}} \sqcap & \exists \text{\textcolor{ForestGreen}{hasPriority}}.1
\end{align}
 \begin{align}
 \label{rule2}
    \text{\textcolor{Mulberry}{Rule2}} \sqsubseteq \text{\textcolor{orange}{ActionPriority}} \sqcap & \exists \text{\textcolor{ForestGreen}{hasAction}.\textcolor{Mulberry}{Put}}
    \nonumber \\
    \sqcap \exists \text{\textcolor{ForestGreen}{hasObjectType}.\textcolor{orange}{FoodItem}} \sqcap & \exists \text{\textcolor{ForestGreen}{hasPriority}}.2
\end{align}
These rules define priority constraints for performing the "Put" action on different object types. Rule 1 establishes that when placing objects, items classified as Crockery should have the highest priority, meaning they should be placed before other object types. This prioritization is explicitly set by assigning a priority value of 1 to Crockery, ensuring that these objects are handled first in the sequence of actions.

Similarly, Rule 2 specifies that objects that belong to the FoodItem category have a lower priority when performing the same "Put" action. With a priority value of 2, FoodItems are placed after Crockery, reflecting a structured ordering in object manipulation. These rules collectively enforce a logical sequence for object handling, ensuring that tasks such as arranging or placing objects follow a predefined hierarchical structure that is suggested to LLM along with the task prompt.

\subsection{Semantic Tagging}\label{sec:tagging}

The proposed system relies significantly on semantic tagging and recognizing tasks and objects from user input.  We use the SpaCy library (\url{https://spacy.io/}), which provides advanced capabilities for part-of-speech (POS) tagging and dependency parsing. It takes user input in the form of natural language and extracts the key \texttt{tasks} (verbs) and \texttt{objects} (nouns) by analyzing the grammatical structure of the sentence.

The user input string is first tokenized into individual words using SpaCy’s tokenization. Each token is annotated with a POS tag that identifies its grammatical role, such as a verb, noun, or preposition. The system iterates over the tokens and identifies the verbs using their POS tag. The task is then extracted by selecting verbs that correspond to a predefined set of valid tasks. We assume that the set of tasks that a robot can perform are: \texttt{tasks = \{"clean", "arrange", "put", "serve", "stack"\}}.

Formally, the input sentence can be represented as a sequence of tokens, i.e., \(\{t_1, t_2, ..., t_n\}\). Each token $t_i$ is assigned a part-of-speech tag $p_i$.\\
Where $p_i \in \{\text{\small VERB}, \text{\small NOUN}, \text{\small ADJ}, ...\}$.  The system identifies a task $\alpha$ such that: 
\(
\alpha = t_i \quad \text{if} \quad p_i = \text{\small VERB} \quad \text{and} \quad t_i \in  \texttt{tasks}.
\)

Once the task is identified, the next step is to extract the objects of the task. We use SpaCy's dependency parsing to identify the direct objects (\texttt{dobj}), prepositional objects (\texttt{pobj}), and conjunct objects (\texttt{conj}). These dependencies are used to capture the relationships between the verbs and their corresponding objects. If $t_j$ represents a token with POS tag \texttt{NOUN}, then $t_j$ is extracted as an object if it satisfies the condition:
\(
\text{Dependency}(t_j) \in \{\texttt{dobj}, \texttt{pobj}, \texttt{conj}\}
\)

In addition to processing single-word objects, the system is designed to handle compound nouns connected by underscores (e.g., \textit{green\_cup}) or adjectives. In the case of compound nouns, the dependency tag \texttt{compound} is used to combine modifiers with the noun. The system ensures that compound words are grouped together, using an underscore (\_) to create a single object label (e.g., \textit{green\_cup}). Formally, if a noun $t_k$ has a compound modifier $t_m$, then the object $O$ is given by:
\(
O = \{t_m\_t_k\}, \; \text{where} \; \text{Dependency}(t_m) = \texttt{compound}
\).


    
    
    
    

\subsection{Contextual Inference Module}\label{sec-inference}
Once the task and the list of objects are identified, as explained in Sec.~\ref{sec:tagging}, the inference module applies a set of SPARQL queries to the ontological knowledge explained in Sec.~\ref{sec:ontology}. These queries are designed to generate textual descriptions that facilitate the LLM in computing a semantically correct task plan. The ontology is represented in RDF (Resource Description Framework) format, which provides a standard way of encoding information about objects, actions, and relationships. Using the rdflib library (\url{https://rdflib.readthedocs.io/en/stable/}), we load the ontology as RDF graphs, which allows us to perform SPARQL queries to access and extract specific knowledge that is essential for the task-planning process. These queries retrieve information about objects, their classifications, and prioritized actions. By executing these queries, the system formulates natural language prompts and then passes to the LLM alongside user input to ensure that the robot executes actions in a logically and semantically accurate sequence.


The first set of SPARQL queries identifies the types of objects referenced in a user command. This classification is essential, as different objects have specific priorities depending on their type and the task at hand. For example, the query below identifies whether an object, such as an ``apple'' or ``plate,'' is categorized as a \textit{FoodItem}, \textit{Crockery}, or \textit{Container}. 

\begin{alltt}
\texttt{\textbf{\textcolor{MidnightBlue}{PREFIX}} ex:} 
\texttt{<http://.../kitchen\_ontology\#>} 
\texttt{\textbf{\textcolor{MidnightBlue}{SELECT}} ?type \textbf{\textcolor{MidnightBlue}{WHERE}} \{ } 
\texttt{\ \ \ \ \ \ \ \ ?obj a ?type ;} 
\texttt{\ \ \ \ \ \ \ \ \textbf{rdfs:label} ?label .}
\texttt{\ \ \ \ \ \ \ \ \textbf{\textcolor{MidnightBlue}{FILTER}} (lcase(str(?label)) = 
\ \ \ \  \ \  \ \  \ \ \ \ \ "{object\_name.lower()}")}
\texttt{\ \ \ \ \ \ \ \ \textbf{\textcolor{MidnightBlue}{FILTER}} (?type != 
\ \ \ \  \ \  \ \  \ \ \ \ \ owl:NamedIndividual)} \}
\end{alltt}

This query is executed by iterating over items identified in the user's input, enabling the system to classify each item and prepare for a task-specific prioritization. The returned object types inform subsequent queries about priorities and relationships, ensuring that the instructions provided to the LLM are contextually accurate.


Once object types are identified, the inference module executes another set of queries to retrieve predefined task descriptions and priorities from the ontology. These descriptions specify the correct order for performing actions based on the object type and task type (e.g., ``put,'' ``clean,'' or ``arrange''). For example, a query to retrieve the priority for any particular action is as follows:

\begin{alltt}
\texttt{\textbf{\textcolor{MidnightBlue}{PREFIX}} ex:} 
\texttt{http://.../kitchen\_ontology\#>}
\texttt{\textbf{\textcolor{MidnightBlue}{SELECT}} ?priority ?description \textbf{\textcolor{MidnightBlue}{WHERE}} \{} 
\texttt{\ \ \ \ \ \ \ ?rule rdf:type \textbf{ex:ActionPriority };} 
\texttt{\ \ \ \ \ \ \ \textbf{ ex:hasAction} "{action}" ;} 
\texttt{\ \ \ \ \ \ \ \ \textbf{ex:hasObjectType} "{object\_type}" ;}
\texttt{\ \ \ \ \ \ \ \ \textbf{ex:hasPriority} ?priority ;} 
\texttt{\ \ \ \ \ \ \ \ \textbf{ex:hasDescription} ?description .}\ \}
\end{alltt}

This query fetches the action priority and a natural language description for each object, allowing the reasoner to compile a comprehensive instruction set. For example, if a \textit{Crockery} item like a ``plate'' has a higher priority than a \textit{FoodItem} like an ``apple'' for the ``put'' action, the system will generate a prompt indicating that the plate should be placed before the apple. This description is adapted to align with the task type, enhancing the LLM's ability to generate a structured and semantically correct plan.



\subsection{Perception and Env. State Descriptor}\label{sec-env}

The perception module and the environment state descriptor are responsible for detecting and identifying objects in the environment and generating a textual description of the scene based on the types of objects and their spatial relationships. The perception module uses a YOLO-based object detection model and NVIDIA’s FoundationPose library, for object recognition and precise pose estimation, respectively. The environment state descriptor applies an ontology-based semantic classification, that allows the system to produce structured, detailed descriptions of the environment, which is useful for understanding the environment state. 

The perception module begins with YOLO, which detects objects in the input image. Each detected object is labeled with a class name, such as ``apple" or ``cup", based on the YOLO that is fine-tuned on YCB dataset, allowing the system to associate each detected object with its specific label.  Once objects are detected and labeled, the same image along with point cloud data is processed by the FoundationPose library to compute the precise 3D pose of each object. 
The environment state descriptor uses the object names and spatial attributes determined by the perception module and applies a set of SPARQL queries on the ontological knowledge to classify each detected object (such as apple and plate as a FoodItem and Crockery) and retrieve relevant geometric information such as object bounding box, object length, and width.

All extracted information is compiled into a textual description that combines both the object’s class type and its computed spatial data. For instance, after processing the object ``apple" the system generates a description indicating the ``apple" is a ``FoodItem" located at position [x,y,z] and orientation [yaw] with a bounding box spanning from [xmin, ymin] to [xmax, ymax] and dimensions of $l$ meters in length and $w$ meters in width. This is done in a similar way as well for all the other detected objects.  This textualized environment state description is then fed to the prompt generator that develops the final prompt for the LLM task planner.

\begin{figure}[t]
	\includegraphics[width=\columnwidth]{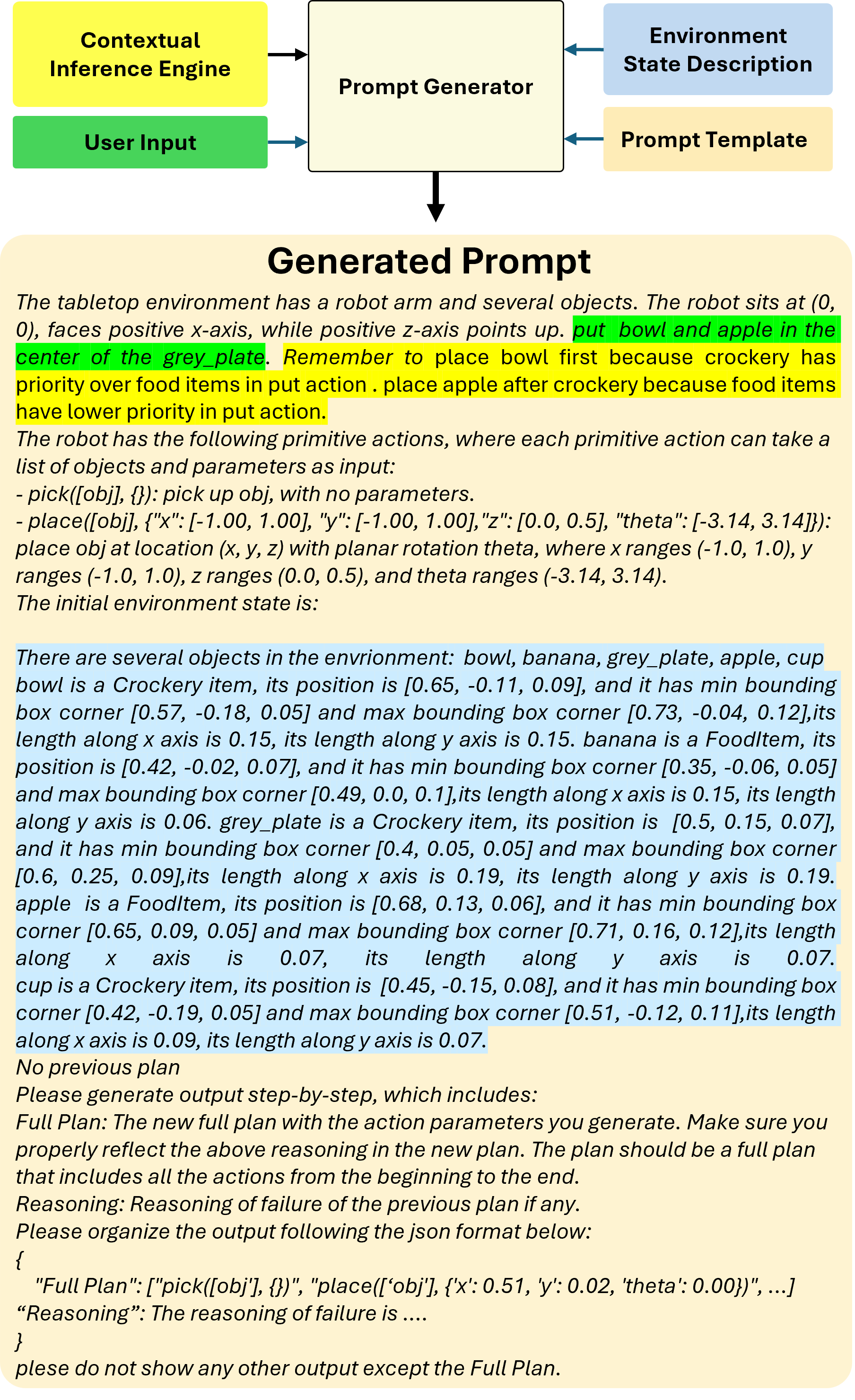}
	\caption{Illustrate how the prompt generator integrates the information of user input (green), contextual inference module (yellow), Environment State Description (blue), and Prompt Template (black), to construct the final system prompt. The Generated Prompt incorporates structured environment data, action constraints, and reasoning to guide robotic decision-making effectively.}
    \vspace{-8mm}
\label{fig:prompt}
\end{figure}

\subsection{Prompt Generator}\label{sec-promptgen}
This module generates the prompt to be sent to the LLM-TAMP module by combining the prompt template with the information coming from the \textit{Contextual Inference Engine} and  \textit{Environmental State Descriptor} modules. For instance, in a user input prompt such as ``\texttt{put the apple and bowl on the tray}", the actual final output prompt feed to the LLM-Task Planner after tuning will be the one shown in Fig.~\ref{fig:prompt} inside the prompt Generated box, where the prompt template is completed with the yellow text that comes from the Contextual Inference Engine, the blue text that comes from the Environment State Descriptor, and the green text that comes from the user input.

\subsection{LLM Task and Motion Planner}\label{sec-taskplanner}
The LLM-Task planner takes a generated prompt as input and formulates a symbolic plan to be executed by the robotic system. The symbolic plan generation closely follows the methodology used in the baseline approach~\cite{Wang2024}. For instance, using the prompt commented in the previous section, the LLM generates the following symbolic plan:

\begin{alltt}
\texttt{\textbf{Full Plan} = }
\texttt{\ \ \ \ \ \ \textbf{\textcolor{MidnightBlue}{Pick}}\ ([\textcolor{Mulberry}{bowl}],\{\}) } 
\texttt{\ \ \  \ \ \textbf{\textcolor{MidnightBlue}{Place}}([\textcolor{Mulberry}{bowl}]),\{x,y,z,\(\theta\)\}}
\texttt{\ \ \ \ \ \ \textbf{\textcolor{MidnightBlue}{Pick}}\ ([\textcolor{Mulberry}{apple}],\{\}) } 
\texttt{ \ \ \ \ \ \textbf{\textcolor{MidnightBlue}{Place}}([\textcolor{Mulberry}{apple}]),\{x,y,z,\(\theta\)\}}
\end{alltt}

\begin{figure*}[t]
	\includegraphics[width=\linewidth]{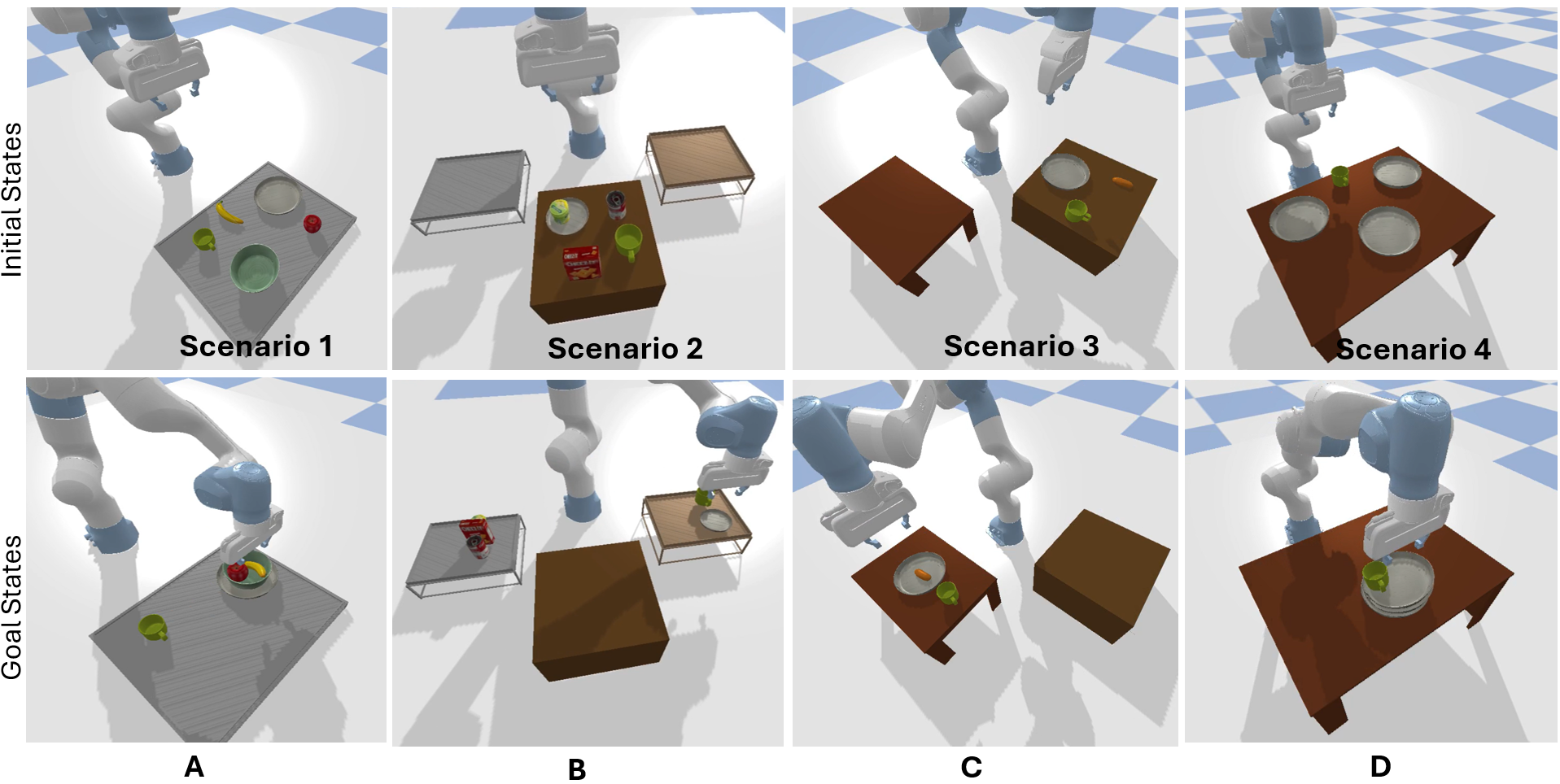}
	\caption{Example scenarios validating the proposed approach: the first row shows the initial states and the second row shows the goal state. each scenario is used to perform multiple tasks, some of them are given below: (A) Task: \textit{Put bowl, banana, and apple on the plate}; (B) Task: \textit{Clean table, move sugar box, tomato can, and cracker box to the left table,
move the plate and cup to the right table}; (C) Task: \textit{Serve breakfast by placing plate, bread, and cup on the table}; (D) Task: \textit{Stack plate1, plate2, and cup on plate3}.} 
\label{fig:scenes}
\end{figure*}

In this example, the generated symbolic plan outlines a sequential set of actions in which the robot system first picks the plate from its initial position and places it at the specified target location with a given orientation. The process is then repeated for the apple, where the planner generates a \texttt{Pick} command followed by a \texttt{Place} command, each with their designated positions and orientations.

The execution of each action in this symbolic plan is managed by a motion planning module that interfaces with a motion planner to compute collision-free paths for the manipulator. For motion planning, we employ RRTConnect, which is well-suited for generating efficient and collision-free trajectories in constrained environments. After each action, the motion planner provides feedback to the LLM-Task Planner, confirming successful execution or detailing any encountered issues. 

In the case of a motion planning failure, e.g. when a target configuration is unreachable or no collision-free path exists, the motion planner will communicate the specific reason for failure to the LLM-Task Planner. Upon receiving this feedback, the LLM-Task Planner adapts by re-evaluating and recalculating the symbolic plan, taking into account the failure conditions. This iterative feedback mechanism ensures robust task execution by enabling the system to dynamically adjust its planning in response to unexpected obstacles or constraints, optimizing both adaptability and reliability in task completion.

\section{Results and Discussion}\label{resultndiscussion}

This section provides a comprehensive evaluation of the performance of the Onto-LLM-TAMP framework and its critical components, including semantic tagging accuracy, effectiveness of knowledge-based reasoning, and comparative benchmarking of various LLM models.
\subsection{Experimental Setup}
\begin{figure}[t]
	\includegraphics[width=\columnwidth]{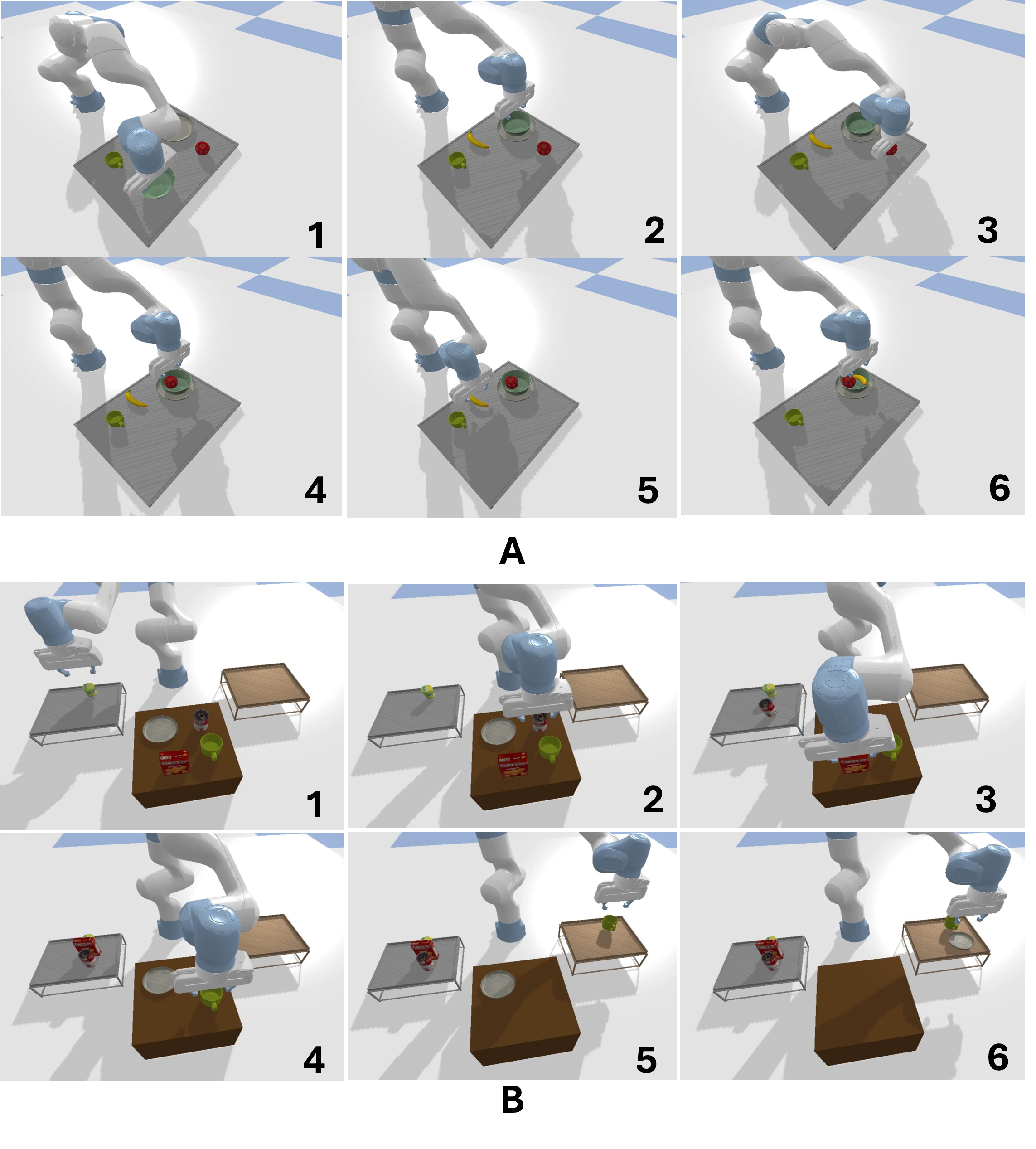}
	\caption{Sequence of snapshots of the following tasks: (A) Task: \textit{Put apple, banana, and bowl in plate}; (B) Task: \textit{Clean table, move plate and cup to the right\_table, move sugar\_box, tomato\_can, and cracker\_box to the left\_table.} }	
\label{fig:screenshot}
\end{figure}


The proposed Onto-LLM-TAMP framework was comprehensively validated through both simulation and real-world experiments, confirming its practical applicability and robustness across diverse task scenarios. Fig.~\ref{fig:scenes} illustrates the simulation scenarios used for validation. These scenarios were implemented using the PyBullet physics engine. Four different task scenarios were designed to thoroughly evaluate the system’s performance. 

In the first scenario (Fig.~\ref{fig:scenes}-A), the task objective involved organizing items by placing an \texttt{apple}, a \texttt{banana}, and a \texttt{bowl} onto a \texttt{plate}. The second scenario (Fig.~\ref{fig:scenes}-B) focused on a table-cleaning task that required the relocation of various objects; specifically, crockery items (\texttt{cup} and \texttt{plate}) had to be moved to the \texttt{left\_table}, whereas boxed food items (\texttt{cracker\_box}, \texttt{sugar\_box}, and \texttt{tomato\_can}) were to be placed on the \texttt{right\_table}. The sequence of execution is depicted in Fig.~\ref{fig:screenshot}.

\begin{figure}
\centering	\includegraphics[width=\columnwidth]{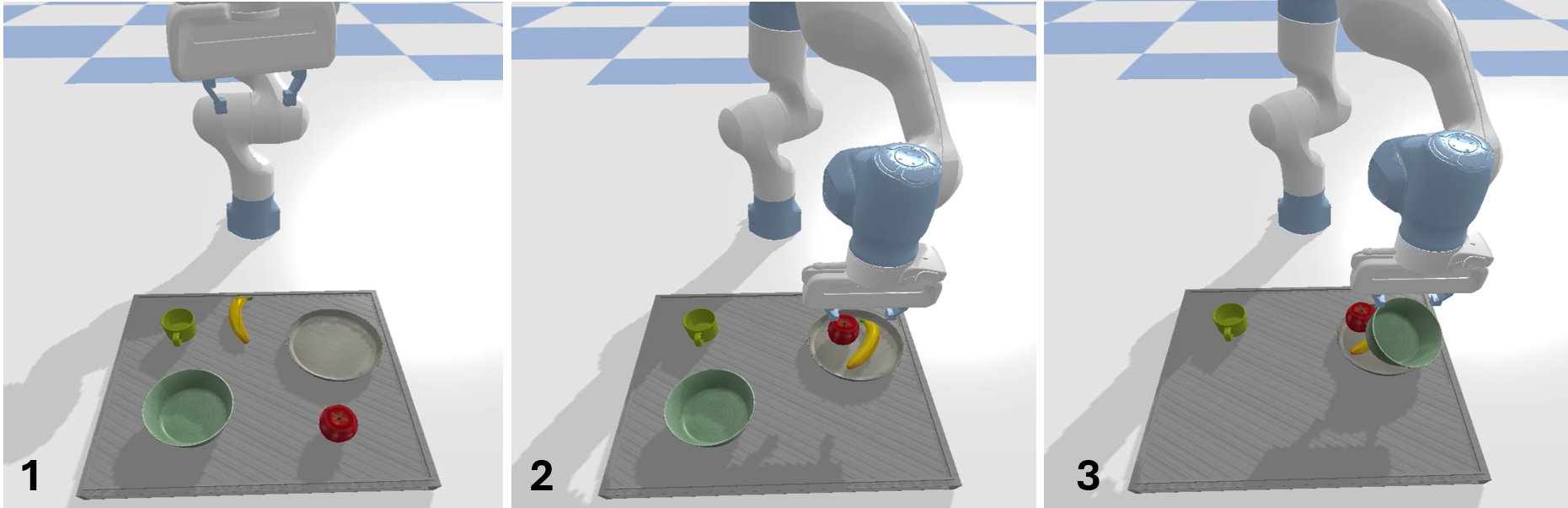}
	\caption{Example of a wrong symbolic plan computation. for the Task: \textit{Put apple, banana, and bowl in plate}. }	\label{fig:wrong}
\end{figure}

The third scenario (Fig.~\ref{fig:scenes}-C) simulated arranging a breakfast table, involves the precise placement of a \texttt{cup} and \texttt{plate} on the table, followed by putting a \texttt{bread} on the plate. The final scenario (Fig.~\ref{fig:scenes}-D) is about stacking objects, where (\texttt{plate1}, \texttt{plate2}, and \texttt{cup}) had to be systematically stacked onto \texttt{plate3}. To further assess the system's robustness, variations in task prompts were introduced by changing the order of objects mentioned by the user.
These scenarios were carefully designed to evaluate the system's capability in managing complex arrangements of objects, where maintaining the correct temporal sequence of subtasks is critical for achieving semantically correct and successful task execution.

\begin{table*}[ht]
\centering
\caption{Comparison of LLM-TAMP and Onto-LLM-TAMP. TPSR: Task planning success rate, EXESR: Execution success rate, CALLs: LLM Calls}
\label{tab:sorted_llm_comparison}
\begin{adjustbox}{max width=\textwidth}
\begin{tabular}{|p{8cm}|c|c|c|c|c|c|}
\hline
\multirow{2}{*}{\textbf{Prompt}} & \multicolumn{3}{c|}{\textbf{LLM-TAMP}} & \multicolumn{3}{c|}{\textbf{Onto-LLM-TAMP}} \\ \cline{2-7} 
                                 & \textbf{TPSR \%} & \textbf{EXESR \%} & \textbf{\# CALLs} & \textbf{TPSR \#} & \textbf{EXESR \#} & \textbf{\# CALLs} \\ \hline
\scriptsize Task 1: Put bowl, banana and apple in plate  & 99.5 & 97.3 & 2.8 & 99.7 & 97.1 & 3.2 \\ \hline
\scriptsize Task 2: Put banana, apple and bowl in plate  & 30.7 & 29.4 & 8.6 & 99.4 & 96.9 & 2.6 \\ \hline
\scriptsize Task 3: Clean table, move sugar\_box, tomato\_can, and cracker\_box to the left table, move plate and cup to the right table  & 96.4 & 89.3 & 7.7 & 98.8 & 93.2 & 6.3 \\ \hline
\scriptsize Task 4: Clean table, move plate and cup to the right table, move sugar\_box, tomato\_can, and cracker\_box to the left table  & 42.8 & 36.6 & 8.4 & 98.3 & 91.4 & 6.8 \\ \hline
\scriptsize Task 5: Clean table, move crockery items to the right table, move boxed food items to the left table & 0 & 0 & 10 & 48.1 & 41.6 & 8.3 \\ \hline
\scriptsize Task 6: Serve breakfast by placing plate, bread and cup on the  table. & 98.4 & 90.3 & 4.6 & 98.7 & 90.1 & 4.9 \\ \hline
\scriptsize Task 7: Serve breakfast by placing bread, plate and cup on the  table. & 20.8 & 19.6 & 9.7 & 98.6 & 92.4 & 4.3 \\ \hline
\scriptsize Task 8: Stack plate1, plate2 and cup on plate3 & 98.9 & 94.8 & 6.7 & 98.7 & 92.1 & 5.3 \\ \hline
\scriptsize Task 9: Stack cup, plate1, and plate2 on plate3  & 17.2 & 12.4 & 9.8 & 98.1 & 93.2 & 5.8 \\ \hline
\scriptsize Task 10: Stack cup and plates on the table & 0 & 0 & 10 & 68.1 & 61.8 & 6.3 \\ \hline
\end{tabular}
\end{adjustbox}
\end{table*}

\subsection{Onto-LLM-TAMP Evaluation}
The above scenarios are used to compare the proposed framework with the baseline approach. The results presented in Table \ref{tab:sorted_llm_comparison} highlight the importance of integrating ontology-based reasoning into LLM-TAMP systems. The parameters we compare are;
\begin{itemize}
\item \textit{Task planning success rate (TPSR)}: Represents the percentage of tasks for which the generated plans by the LLM correctly capture the intended task semantics and objectives, resulting in an executable action sequence.
\item \textit{Execution success rate (EXESR)}: Indicates the percentage of tasks successfully completed when executing the generated plans on the robot, highlighting the practical feasibility and effectiveness of the plans.
\item \textit{Number of LLM calls (\# CALLs)}: Measures how many LLM calls were required to generate a suitable action plan, reflecting the efficiency and complexity involved in the planning process.
\end{itemize}
The Onto-LLM-TAMP consistently outperforms the baseline LLM-TAMP, particularly in complex scenarios that require refined reasoning and object categorization.
For straightforward tasks with explicit and correctly ordered object descriptions, such as \textit{Task 1: Put bowl, banana, and apple in plate}, both Onto-LLM-TAMP and traditional LLM-TAMP systems achieve nearly perfect performance (approximately 99.5\% TPSR and above 97\% EXESR), requiring only a minimal number of LLM calls. However, when the order of objects is altered, as seen in \textit{Task 2}, the traditional LLM-TAMP's performance significantly drops to about 30.7\% TPSR and 29.4\% EXESR, along with an increased number of LLM calls (8.6 on average). In contrast, Onto-LLM-TAMP maintains high performance (99.4\% TPSR and 96.9\% EXESR) with fewer LLM calls (2.6 on average). This robustness improvement for the Onto-LLM-TAMP is due to the explicit incorporation of the correct semantic sequence of actions using knowledge-based reasoning, which reduces ambiguity in the task. The reason for the failure of LLM-TAMP is that in most of the cases, it ended up performing the task as shown in Fig.~\ref{fig:wrong}. 

\begin{figure*}[t]
\centering
\includegraphics[width=\linewidth]{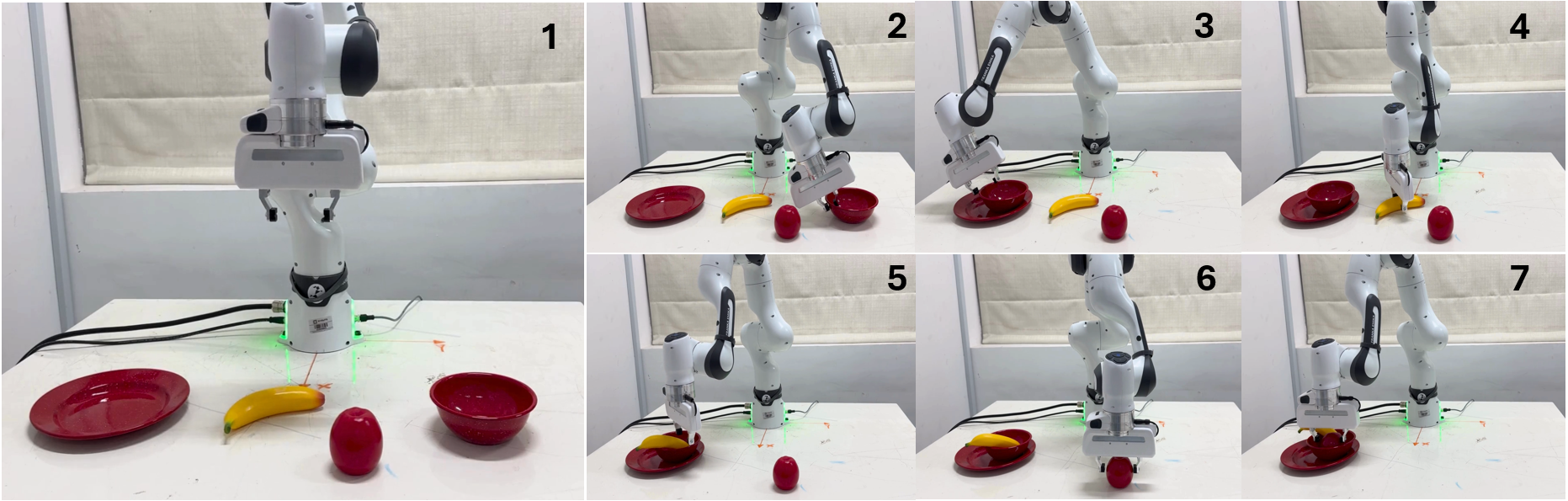}
	\caption{Screenshots of the execution in the real environment. The task was to \textit{put banana, apple and bowl into the plate}  }	\label{fig:real}
\end{figure*}

In more complex scenarios, such as \textit{Task 3}, which involves detailed instructions about moving various items (crockery and boxed foods) to specific locations, both approaches initially perform similarly with high success rates (over 96\% TPSR and around 90\% EXESR). However, when the task description sequence is reversed and made ambiguous (as in \textit{Task 4} and \textit{Task 5}), the baseline approach suffers a significant performance decline, reaching as low as 0\% in TPSR and EXESR. The Onto-LLM-TAMP, meanwhile, maintains a consistently high-performance level (TPSR around 98\% and EXESR around 91\% for \textit{Task 4}, and modest success of approximately 48\% TPSR for the most ambiguous \textit{Task 5}). This capability results from using structured ontological knowledge to precisely categorize items (e.g., identifying \texttt{cracker\_box} as boxed food and \texttt{plate} as crockery), thus providing the LLM with clearer contextual information.

For tasks that involved serving and stacking items (\textit{Tasks} 6 to 10), the Onto-LLM-TAMP consistently demonstrates superior performance compared to the baseline. In \textit{Tasks 6 and 7}, which involve serving breakfast, the baseline's TPSR drops to around 20\% when the item order is changed. In contrast, the Onto-LLM-TAMP maintains consistently high success rates (above 98\% TPSR and around 92\% EXESR). Similarly, in stacking tasks (\textit{Tasks 8 and 9}), the Onto-LLM-TAMP achieves high TPSR (around 98\%) and EXESR (above 92\%), significantly outperforming the baseline, especially when the item order is varied.
\textit{Task 10}, involving stacking items ambiguously described as \textit{cup and plates on the table}, highlights the limits of both approaches; the Onto-LLM-TAMP approach still achieves success (68.1\% TPSR and 61.8\% EXESR) because the knowledge reasoner described: \textit{the objects with the large bounding boxes will come first in stacking}. These results indicate the Onto-LLM-TAMP approach’s capability to handle ambiguity and complex semantic scenarios more effectively.

\begin{figure}[t]
	\includegraphics[width=\columnwidth]{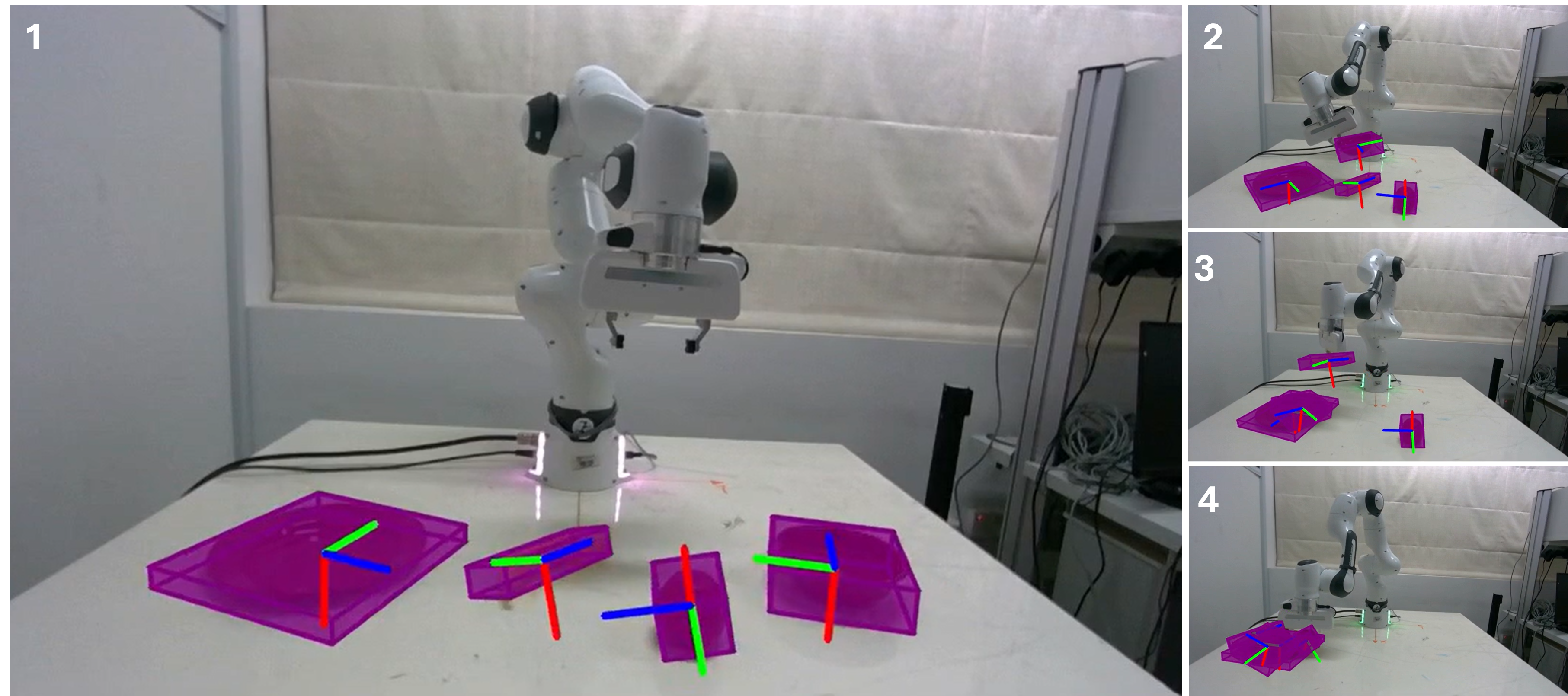}
	\caption{Screenshots of pose estimation during the execution in the real environment. The task was to put banana, apple, and bowl into the plate. }	
\label{fig:pose}
\end{figure}
\begin{figure}[t]
	\includegraphics[width=\columnwidth]{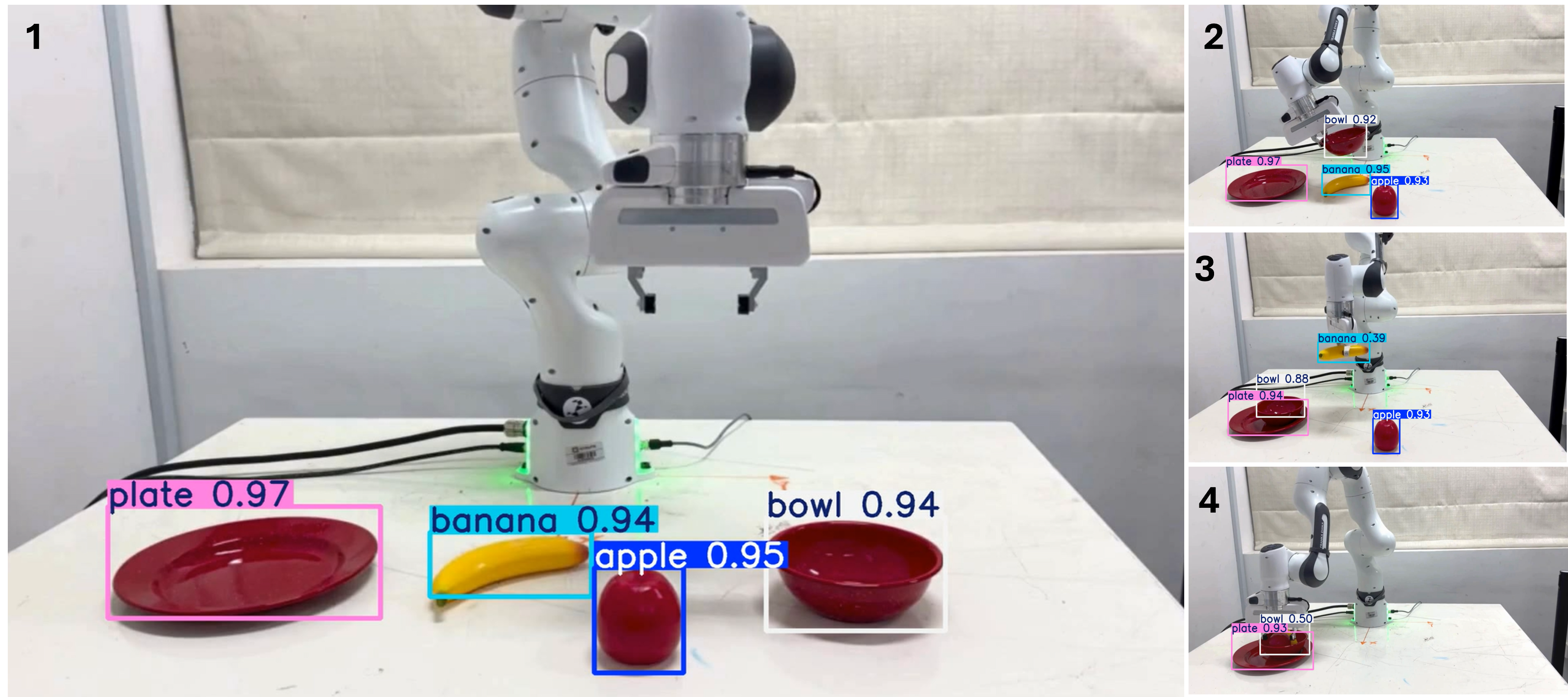}
	\caption{Screenshots of object identification during the execution in the real environment. The task was to put banana, apple, and bowl into the plate. }	
\label{fig:yolo}
\end{figure}
During the real-world demonstration, we execute the task of \textit{put a banana, an apple, and a bowl in the plate}, showcasing the system's ability to accurately detect, localize, and manipulate objects in a real environment. Figure~\ref{fig:real} presents a series of screenshots capturing key moments of the execution process.
To ensure precise object identification and spatial positioning, we utilize YOLO-V8 for object detection and FoundationPose for pose estimation. YOLO-V8 enables real-time recognition of objects in the scene, while FoundationPose estimates their positions and orientations, allowing the robot to plan and execute accurate pick-and-place actions. Figures~\ref{fig:yolo} and~\ref{fig:pose} illustrate the intermediate steps, including detected objects and their estimated poses, demonstrating the effectiveness of our approach in real-world conditions.

\subsection{Semantic Tagging and Reasoning Evaluation}

\begin{table}[t]
    \centering
    \caption{Processing time and accuracy of semantic tagging (ST) and reasoning process (RP) while tuning the prompts.}
    \resizebox{\columnwidth}{!}{
    \begin{tabular}{c|cc|cc}
        \toprule
        \multirow{2}{*}{\scriptsize{Experiment}} & \multicolumn{2}{c|}{Time (s)} & \multicolumn{2}{c}{Accuracy (\%)} \\
        & ST & RP & ST & RP  \\
        \midrule
        \scriptsize{Exp 1 (Task 1-10)}  & 0.0048 & 0.094 & 98.43 & 89.71 \\
       \scriptsize{ Exp 2 (Random Tasks)}  & 0.0071 & 0.109 & 81.51 & 77.94 \\
        \bottomrule
    \end{tabular}}
    \label{tab:st_reasoning}
\end{table}
Semantic tagging and knowledge-based reasoning are two core components of the proposed framework. Semantic tagging analyzes the input prompt to identify the intended action and associated objects. Subsequently, the knowledge-based reasoner utilizes these extracted actions and objects to enrich and clarify the input, resulting in a more precise and effective task description.  Table \ref{tab:st_reasoning} presents two distinct measures for evaluating the performance of semantic tagging and knowledge-based reasoning.

\begin{itemize}
\item  \textit{Time:} It quantifies how long it takes for semantic tagging to extract actions and objects, as well as how long the ontology-based reasoner needs to process those extracted elements. 
\item \textit{Performance accuracy:} Indicates how precisely each component performs its function: For semantic tagging, the percentage reflects the accuracy with which actions and objects are identified in the input prompt. For reasoning, it shows how accurately the ontology-based approach infers relationships and enhances the user prompt.
\end{itemize}
Table~\ref {tab:st_reasoning} uses these measures to highlight the speed and precision of the two main processing steps. Overall, the results show that semantic tagging is consistently fast yet experiences a drop in accuracy when dealing with less structured input. Meanwhile, reasoning involves slight computational overhead to perform ontology-driven inferences and similarly shows fairly good accuracy when dealing with a more diverse language, reflecting the impact of varying input complexity.

In experiment 1, we applied semantic tagging and reasoning on each task 10 times and computed the average time and accuracy for 10 tasks. Semantic tagging requires an average of 0.0048 seconds to process the input string and achieve an accuracy of 98.43\%. This suggests that for these structured tasks, the tagging process is efficient and highly accurate in recognizing relevant actions and objects. Meanwhile, the reasoning step reports an average processing time of 0.094 seconds with an accuracy of 89.71\%. This indicates that once actions and objects are identified, the reasoner uses the ontology to derive correct inferences and enhance the input prompt most of the time. However, a small fraction of tasks may involve entity mismatches or incomplete definitions.

In experiment 2, we used random task prompts. Comparable to tasks 1-10, but introduced some typos, misplaced commas, and other parts of speech. The semantic tagging time increases slightly to 0.0071 seconds. It reflects that tasks expressed in less predictable language can require slightly more processing. The corresponding accuracy of 81.51\% shows that the NLP component finds the correct actions and objects in most but not all cases due to more variable and ambiguous language. The reasoner’s processing time is 0.109 seconds, indicating that it can handle these more diverse inputs but does so with a bit more computational load. Its accuracy of 77.94\% suggests that knowledge-based inference remains generally effective but can occasionally fail to match extracted entities to the ontology when the input is unfamiliar.

\begin{table*}[ht]
    \centering
    \caption{"Comparison of LLM models on Tasks 1 to 10 for LLM-TAMP and Onto-LLM-TAMP, evaluated based on two key metrics: execution time (T) in seconds and correctness (C) score in \%. }
    \label{tab:bleu_score}
    \resizebox{\textwidth}{!}{%
    \begin{tabular}{c|cccc|cccc|cccc|cccc}
        \toprule
        \multirow{2}{*}{\textbf{Task}} 
        & \multicolumn{4}{c|}{\textbf{GPT-4}} 
        & \multicolumn{4}{c|}{\textbf{Gemini}} 
        & \multicolumn{4}{c|}{\textbf{Cohere}} 
        & \multicolumn{4}{c}{\textbf{LLaMA}} \\
        \cmidrule(lr){2-17}
        & \multicolumn{2}{c}{LLM-TAMP} & \multicolumn{2}{c|}{Onto-LLM-TAMP} 
        & \multicolumn{2}{c}{LLM-TAMP} & \multicolumn{2}{c|}{Onto-LLM-TAMP} 
        & \multicolumn{2}{c}{LLM-TAMP} & \multicolumn{2}{c|}{Onto-LLM-TAMP} 
        & \multicolumn{2}{c}{LLM-TAMP} & \multicolumn{2}{c}{Onto-LLM-TAMP} \\
        \cmidrule(lr){2-17}
        & T & C & T & C
        & T & C & T & C
        & T & C & T & C
        & T & C & T & C \\
        \midrule
        1  & 2.55 & 100 & 2.62 & 100 & 2.31 & 100 & 2.15 & 100 & 3.01 & 100 & 3.03 & 100 & 3.92 & 100 & 3.98 & 100 \\
        2  & 3.25 & 0   & 3.61 & 100 & 2.10 & 0   & 2.41 & 100 & 3.02 & 10 & 2.44 & 100 & 3.57 & 0   & 3.32 & 100 \\
        3  & 3.19 & 100 & 4.87 & 100 & 3.47 & 100 & 2.96 & 100 & 3.07 & 30  & 3.35 & 30  & 4.91 & 90  & 5.28 & 90  \\
        4  & 2.86 & 10  & 3.66 & 100 & 3.53 & 10  & 3.48 & 100 & 3.06 & 0   & 3.49 & 0   & 3.75 & 10  & 2.89 & 100 \\
        5  & 3.12 & 10  & 3.80 & 100 & 3.53 & 10  & 3.48 & 100 & 3.61 & 0   & 2.90 & 0   & 4.61 & 10  & 4.96 & 50  \\
        6  & 3.23 & 100 & 3.72 & 100 & 1.99 & 100 & 2.71 & 100 & 3.75 & 90  & 3.24 & 90  & 4.75 & 90  & 4.45 & 90  \\
        7  & 2.60 & 10  & 3.70 & 100 & 2.68 & 10  & 2.74 & 100 & 4.23 & 10  & 3.31 & 90  & 5.09 & 0   & 4.95 & 90  \\
        8  & 3.60 & 100 & 3.62 & 100 & 2.93 & 100 & 2.06 & 10  & 3.18 & 100 & 3.92 & 100 & 3.22 & 10  & 4.91 & 10  \\
        9  & 3.82 & 10  & 3.72 & 100 & 2.50 & 10  & 2.86 & 10  & 4.01 & 10  & 3.28 & 10  & 4.03 & 10  & 4.23 & 10  \\
        10 & 4.29 & 10  & 4.65 & 10  & 2.86 & 10  & 2.74 & 10  & 3.35 & 10  & 3.12 & 10  & 4.04 & 10  & 4.80 & 100 \\
        \midrule
        \textbf{Average} & 
        \textbf{3.25} & \textbf{55.0} & \textbf{3.80} & \textbf{91.0} & 
        \textbf{2.79} & \textbf{55.0} & \textbf{2.76} & \textbf{83.0} & 
        \textbf{3.43} & \textbf{36.0} & \textbf{3.21} & \textbf{63.0} & 
        \textbf{4.19} & \textbf{34.0} & \textbf{4.28} & \textbf{74.0} \\
        \bottomrule
        
    \end{tabular}%
    }
\end{table*}

\subsection{LLM Models Evaluations}
The quality and correctness of the computed symbolic plan largely depend on the clarity of the prompt and the capabilities of the LLM model. To evaluate this, we assessed the performance of various LLM models, specifically GPT, LLaMA, Gemini, and Cohere in both standard LLM-TAMP and onto-LLM-TAMP frameworks. The results of this evaluation are summarized in Table~\ref{tab:bleu_score}. The experiments measure two key parameters for both approaches: 
\begin{itemize}

\item \textit{Time:} Defined as the duration each LLM requires to generate executable task plans using API call. 
\item \textit{Correctness:}, Assessed by two robotics experts who manually evaluated each generated action sequence. The evaluation criteria combined semantic accuracy (90\%) and structural correctness (10\%). High correctness scores indicate that the generated plans are not only syntactically executable but also semantically coherent and realistically applicable to robotic scenarios.
     
\end{itemize}
\begin{figure}[t]
	\centering
\includegraphics[width=\columnwidth]{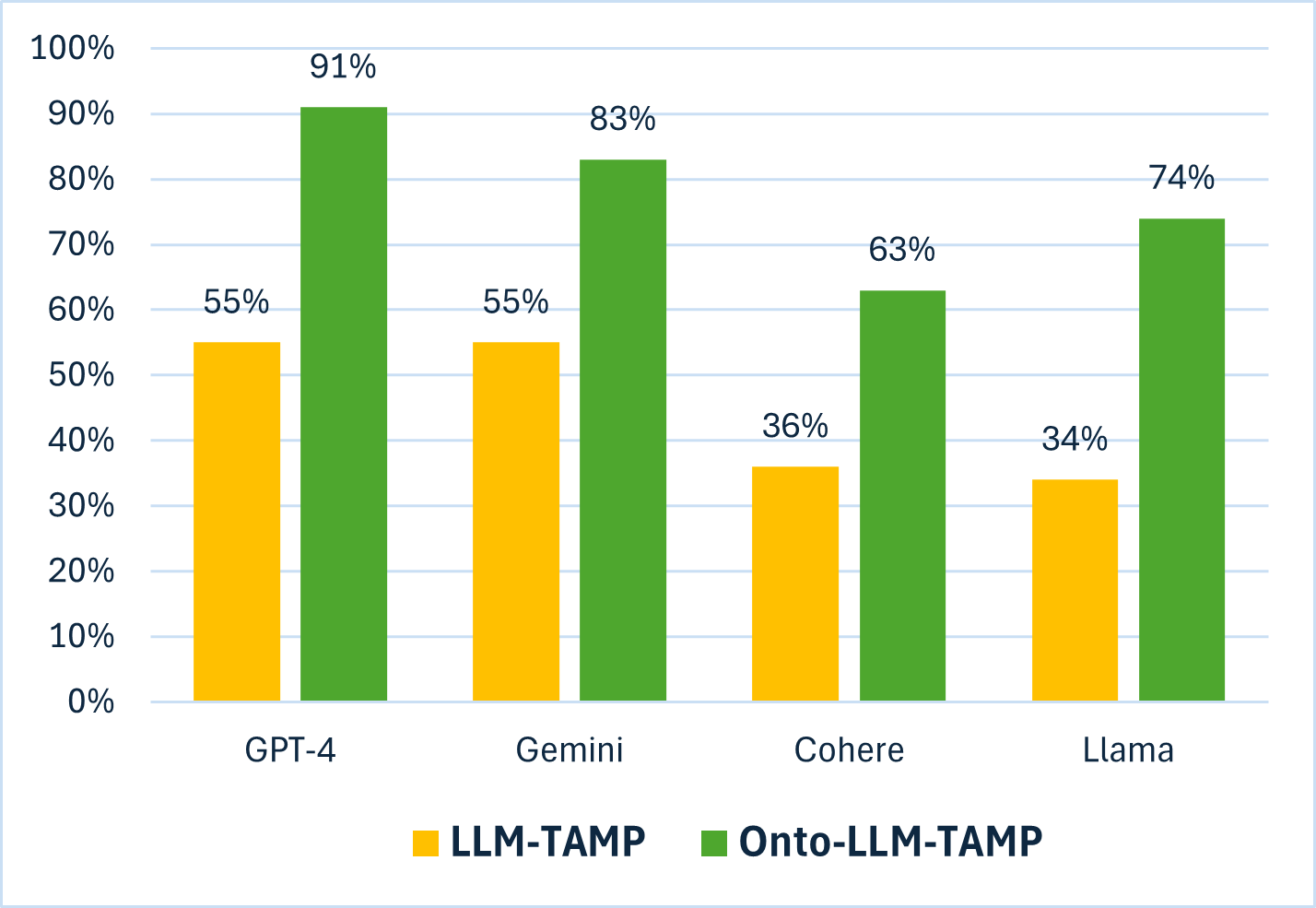}
\caption{Histogram showing the average correctness scores of LLM models for LLM-TAMP and Onto-LLM-TAMP cases.}\label{fig:histogram}
\end{figure}

For tasks with straightforward, clearly structured descriptions (e.g., Task 1 and Task 8), both methods performed comparably well, achieving high correctness scores (ranging from approximately 92\% to 100\%). This consistency demonstrates that, when task instructions are explicit and sequentially correct, LLM-based approaches reliably translate user prompts into executable task plans. Time performance in such tasks varied slightly while generally remaining stable across models and methods. 
However, when tasks contained ambiguity, non-explicit ordering, or object arrangements different from the correct execution sequence (e.g., Task 2: "Put banana, apple, and bowl in plate," Task 4, and Task 5), the performance of the LLM models significantly declined. It can be observed by correctness scores that drop sharply (often to 0–10\%) when the user task prompt is directly provided as input to the LLM model (LLM-TAMP). Such tasks posed a major challenge for LLM-TAMP due to the inherent ambiguity or lack of structured context in the provided input, leading to an unreliable sequence of actions. In contrast, when the user prompts along with the ontological reasoning passed to the LLM Models (onto-LLM-TAMP), it consistently maintained high correctness (usually around 100\%), clearly showcasing its ability to handle ambiguity through ontological knowledge integration effectively. By utilizing structured domain knowledge of onto-LLM-TAMP, the LLM models were able to interpret ambiguous instructions more precisely, ensuring semantically coherent action sequences.

Interestingly, even though onto-LLM-TAMP involves an additional reasoning step that potentially increases the overall processing complexity, it still achieves comparable execution times in most of the cases. This indicates that ontology-driven clarification of task semantics can streamline the ensuing LLM processing by providing clearer input.
\begin{figure}[t]
	\centering
\includegraphics[width=\columnwidth]{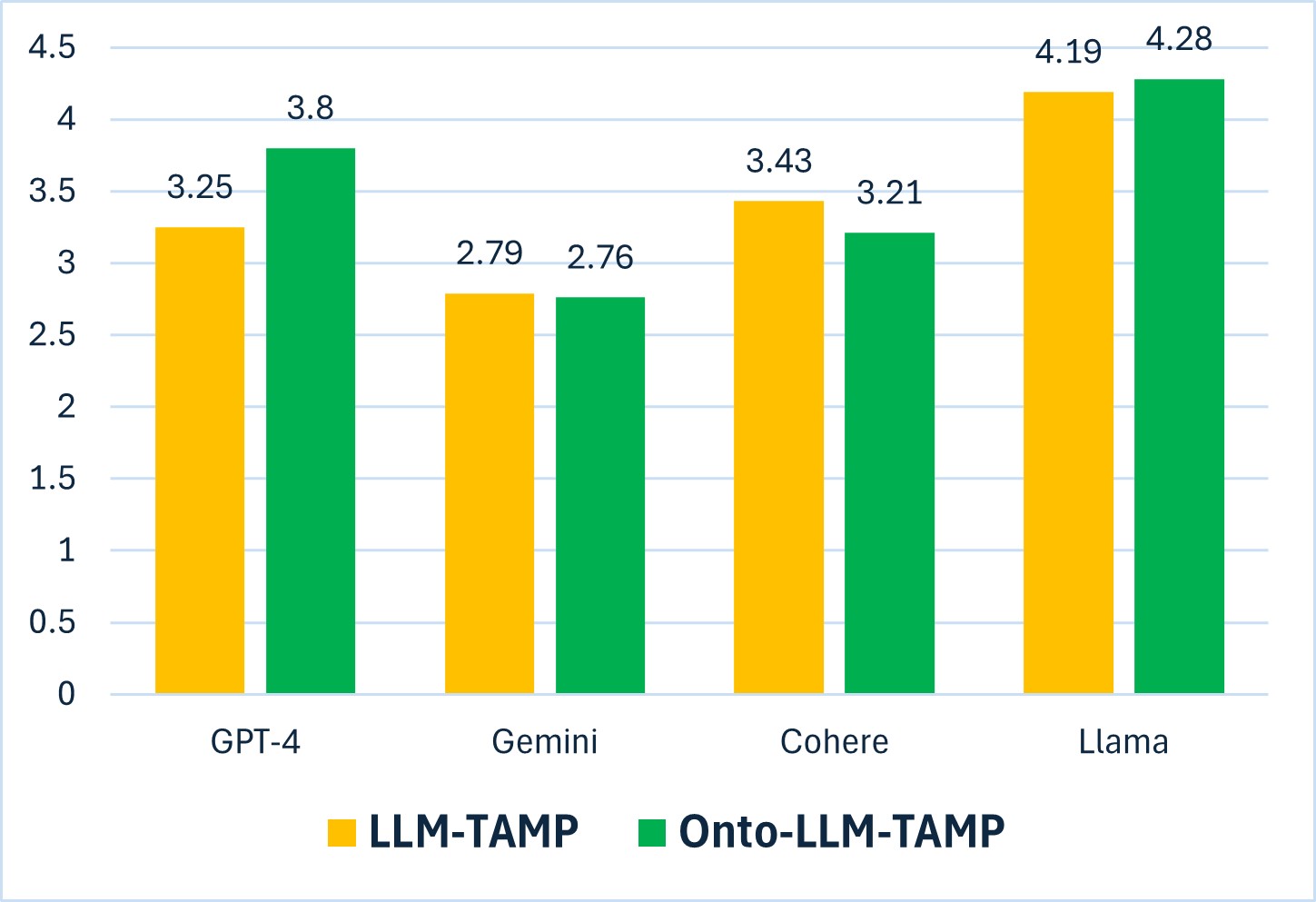}
\caption{Histogram showing the average API call time of LLM models for LLM-TAMP and Onto-LLM-TAMP cases.}\label{fig:histogramtime}
\end{figure}

The average results presented at the end of Table \ref{tab:bleu_score} clearly illustrate the overall advantages of employing an Onto-LLM-TAMP approach compared to the standard LLM-TAMP setting across all evaluated models. Specifically, the average correctness improved significantly when using the ontological enhancement, reflecting enhanced semantic understanding and reduced ambiguity in generated action sequences. For clarity, we have added a histogram showing the average correctness score of LLM models for the cases of LLM-TAMP and Onto-LLM-TAMP. In correctness,  GPT-4 showed an improvement from 55\% correctness in standard LLM-TAMP to 91\% correctness with ontology integration. Similar trends are observed for Gemini (from 55\% to 83\%), Cohere (from 45\% to 63\%), and LLaMA (from 34\% to 74\%).

In terms of computational efficiency, measured by the average API calls response time, the Onto-LLM-TAMP maintained comparable response times compared to the basic LLM-TAMP method. Fig.~\ref{fig:histogramtime} shows the histogram of the comparison of average API call time of LLM models in LLM-TAMP and Onto-LLM-TAMP. For GPT-4 and LLaMA, the Onto-LLM-TAMP required a modest increase in average time (3.25s to 3.80s and 4.19s to 4.28s, respectively). Conversely, Gemini and Cohere remained stable (Gemini: 2.79s to 2.76s; Cohere: 3.43s to 3.21s). These variations indicate that while ontology integration slightly affects response times, the substantial improvements in correctness significantly surpass minor time trade-offs. 

\subsection{Motion Planning Time Evaluation}

 \begin{table}[t]
    \centering
    \caption{Average Motion Planning Time (seconds) for Tasks 1–10, Comparing LLM-TAMP and Onto-LLM-TAMP.}
    \begin{tabular}{c|cc}
        \toprule
        \multirow{2}{*}{Task} & \multicolumn{2}{c}{Motion Planning Time (s)} \\
        & LLM-TAMP & Onto-LLM-TAMP \\
        \midrule
        Task 1  & 0.13 & 0.11 \\
        Task 2  & 2.62 & 0.14 \\
        Task 3  & 0.61 & 0.60 \\
        Task 4  & 6.48 & 0.93 \\
        Task 5  & NA & 3.72 \\
        Task 6  & 0.33 & 0.28 \\
        Task 7  & 1.89 & 0.30 \\
        Task 8  & 0.27 & 0.25 \\
        Task 9  & 4.36 & 0.23 \\
        Task 10 & NA & 1.59 \\
        \bottomrule
    \end{tabular}
    \label{tab:motion_planning_time}
\end{table}

Motion planning is another critical component of the proposed framework. Its performance significantly influences the overall efficiency of task execution. To evaluate this aspect, we compared the motion planning performance of the standard LLM-TAMP and the Onto-LLM-TAMP approaches using the RRTConnect algorithm. The results are summarized in Table~\ref{tab:motion_planning_time}. Motion planning times were calculated as the average planning time of multiple calls of the motion planning during the execution of the subtask of each given task. 

For relatively simple tasks (Tasks 1, 3, 6, and 8), both approaches exhibited comparable planning times, indicating the minimal impact of ontological reasoning when task instructions are clear and straightforward. For instance, Task 1 demonstrated similar results (0.13s vs. 0.11s), while slightly more complex tasks, like Task 3, showed nearly identical performance (0.61s vs. 0.60s). However, even moderately complex tasks with slight ambiguity, such as Task 2 and Task 7, significantly benefited from ontology-driven reasoning, resulting in notably reduced planning times (from 2.62s to 0.14s and 1.89s to 0.30s, respectively). 
Ontology-driven reasoning showed significant advantages in complex or ambiguous scenarios (Tasks 4, 5, 9, and 10). For example, in the case of Task 4, planning times were reduced from 6.48s to 0.93s. Tasks 5 and 10,  completely failed under standard LLM-TAMP, and succeeded efficiently (3.72 and 1.59s, respectively) with the Onto-LLM-TAMP.

The primary reason for the significant reduction in planning time observed with the Onto-LLM-TAMP is that, in scenarios where task instructions or object ordering are ambiguous or incorrect, the conventional LLM-TAMP method often struggles to find collision-free trajectories. This struggle causes repeated failures at the motion planning stage, which then prompts the LLM planner to recompute the symbolic plan continuously, consequently increasing the total motion planning duration, particularly in complex tasks.

\section{Conclusions}\label{conclusion}
This work introduced a novel Onto-LLM-TAMP framework to enhance task and motion planning with LLMs. Integrating knowledge-based reasoning and ontology-driven environment state descriptions, the framework dynamically refines user prompts to generate semantically accurate and context-aware symbolic plans. Unlike static, template-based approaches, it addresses flaws such as incorrect temporal goal ordering and adapts effectively to dynamic environments. Validation in both simulation and real-world scenarios demonstrated significant improvements in planning accuracy and adaptability, highlighting the potential of combining LLMs with ontological reasoning for advanced robotic planning in complex tasks and dynamic contexts.

\section*{Declaration}
During the preparation of this work, the author (s) used ChatGPT and Gemini to improve language and readability. After using this tool/service, the author(s) reviewed and edited the content as needed and takes (s) full responsibility for the content of the publication.
\balance

\bibliographystyle{ieeetr}
\bibliography{refs}

\end{document}